\documentclass[11pt]{article}
\usepackage{acl2015}
\usepackage{times}
\usepackage{url}
\usepackage{latexsym}

\usepackage{booktabs} 
\usepackage{graphicx}
\usepackage{multirow}
\usepackage{makecell}
\usepackage{amsmath}
\usepackage{ragged2e}
\usepackage{amssymb}
\usepackage{multirow}

\title{Cross-lingual Word Sense Disambiguation using mBERT Embeddings with Syntactic Dependencies}

\author{Zhu Xingran \\
  Department of Linguistics and Philology \\
  Uppsala University \\
  {\tt Xingran.Zhu.2781@student.uu.se} \\}

\date{}

\begin{document}
\maketitle
\begin{abstract}
Cross-lingual word sense disambiguation (WSD) tackles the challenge of disambiguating ambiguous words across languages given context. The pre-trained BERT embedding model  has been proven to be effective in extracting contextual information of words, and have been incorporated as features into many state-of-the-art WSD systems. In order to investigate how syntactic information can be added into the BERT embeddings to result in both semantics- and syntax-incorporated word embeddings, this project proposes the concatenated embeddings by producing dependency parse tress and encoding the relative relationships of words into the input embeddings. Two methods are also proposed to reduce the size of the concatenated embeddings. The experimental results show that the high dimensionality of the syntax-incorporated embeddings constitute an obstacle for the classification task, which needs to be further addressed in future studies.
\end{abstract}

\section{Introduction}
\textbf{Polysemy} and \textbf{homonymy} are two semantic phenomena that commonly occur in our daily use of language. A polysemy is a word that possesses several separate meanings which are different but \emph{related} to one another. For example, the English verb 'get' is polysemous as it can mean 'procure', 'understand' or 'become'. Homonyms are two words that are spelled and pronounced the same but have different and \emph{unrelated} meanings. For example, the English verb 'bark' meaning the sound made by a dog and the English noun 'bark' meaning the out layer of a tree are homonymous to each other. Because both polysemy and homonymy are words that have the same spelling but different (either related or unrelated) meanings, they are referred to as \textbf{ambiguous words} in this project.\\
\textbf{Word sense disambiguation} (\textbf{WSD}) is the task of identifying the exact meaning of an ambiguous word given context. Existing WSD approaches can be categorised into three main categories: knowledge-based, supervised, and unsupervised \cite{10.1145/1459352.1459355}. \textbf{Knowledge-based} WSD methods rely on different knowledge sources, and a typical knowledge-based approach links latent representations of ambiguous words to an explicit sense inventory such as WordNet \cite{WordNet}. \textbf{Supervised} WSD methods, usually not relying on any external knowledge source, focus on using machine learning algorithms and manually designed features to train a WSD classifier on manually created sense-annotated data. \textbf{Unsupervised} WSD methods generally don't assign explicit senses to ambiguous words but instead learn to discriminate them in un-annotated corpora. Studies comparing these three types of WSD methods have shown that currently the best performing WSD systems are those based on \textbf{supervised} learning, as attested in public evaluation exercises \cite{10.5555/1661445.1661686,snyder-palmer-2004-english,pradhan-etal-2007-semeval}. For example, in the SENSEVAL-2 English all-words task, the supervised systems achieved the best overall performance \cite{Palmer2001EnglishTA}, and in the evaluation exercise by Benjamin Snyder and Martha Palmer, all seven supervised systems had higher precision and recall scores than all nine unsupervised systems \cite{snyder-palmer-2004-english}. The SENSEVAL and SemEval tasks provide a lot of valuable manually annotated data that can be used for training a supervised WSD classifier, and more and more studies in WSD begin to focus on improving the current supervised WSD systems by leveraging different sources of knowledge.\\
Moreover, the WSD tasks can sometimes cover multiple languages, where the target ambiguous words are translations of one another. For example, the ongoing SemEval-2021 Task 2 tackles the challenge of analysing ambiguous words in a cross-lingual setting. One typical sentence pair in the task contains the English sentence 'Click the right \emph{mouse} button' and the French sentence 'Le chat court après la \emph{souris}', where \emph{mouse} and \emph{souris} are translations of each other but have different meanings in the respective sentences.\\
Therefore, the main purpose of this project is to design a \textbf{supervised cross-lingual WSD system} that incorporates two sources of knowledge: the \textbf{pre-trained BERT embeddings} and \textbf{dependency parsing}. The hope is to investigate the WSD performance compared with the current baseline systems, and shed light on the potential of incorporating syntactic information into WSD classification models.

\section{Related Work}
\textbf{BERT} \cite{devlin-etal-2019-bert}, a state-of-the-art language representation model proposed by researchers at Google AI Language in 2018, have been designed to pre-train deep representations of words based on large unlabelled corpora. The BERT model has been proven to be effective in extracting word features and contextual information from plain text, and therefore many studies have tried to incorporate the pre-trained BERT embeddings as features into WSD systems \cite{huang-etal-2019-glossbert,du2019using}.\\
The contextualised BERT embeddings have been shown to be capable of clustering polysemic words into distinct sense regions in the embedding space \cite{Wiedemann2019DoesBM}. Huang et al. from Fudan University leveraged gloss knowledge and proposed three BERT-based models for WSD. They constructed context-gloss pairs and converted WSD to a sentence-pair classification task, based on which they fine-tuned the pre-trained BERT model and achieved state-of-the-art results on WSD task \cite{huang-etal-2019-glossbert}. Similarly, Du et al. from Tsinghua University utilised sense definitions to train a unified WSD classifier and explore several ways of combining BERT and the classifier. In particular, they fine-tuned the BERT on WSD task, and used pre-defined sense definitions to address unseen ambiguous words. Their study showed that a WSD system using pre-trained language models can further benefit from incorporating external knowledge \cite{du2019using}.\\
Although WSD tasks mainly address the semantic aspect of words, the \textbf{syntactic} aspect of words can also be useful for revealing their contextual semantic information. \textbf{Dependency parsing} is the task of analysing the syntactic structure of a sentence and establishing syntactic relationships between \emph{head} words and their modifiers (i.e. \emph{dependent} words) \cite{10.5555/555733}. For example, in the sentence 'The cat chases the mouse', the word 'chases' is the head word of 'mouse', and in the sentence 'Click the right mouse button', 'mouse' has the word 'button' as its head. Given the same word 'mouse' that appears in different contexts with different meanings, taking into consideration the information about its different head words might help to better analyse its semantic meanings in these different contexts.\\
Studies have shown that word embeddings can be enhanced by incorporating syntactic information. In their study, Xu et al. investigated the incorporation of syntactic dependencies to enhance semantic representations of word embeddings to address the issue of different syntactics across languages. Their Bilingual Word Embedding integrating syntactic Dependencies (DepBiWE) achieved state-of-the-art results on various NLP tasks \cite{ijcai2018-628}.

\section{Data Preparation}
This project uses the data from the ongoing SemEval-2021 Task 2\footnote{\url{https://competitions.codalab.org/competitions/27054}}, and the basic information about the data can be found in Table \ref{tab:table1}\footnote{The test set has not been released.}.
\begin{table*}[t]
\centering
\begin{tabular}{l l l l}
\toprule
\textbf{Data} & \textbf{No. of Sentences} & \textbf{No. of Sentence Pairs} & \textbf{Language} \\
\midrule[2pt]
\textbf{Training} & 12,000 & 8,000 & English\\[3pt]
\midrule
\textbf{Development} & 1,500 & 1,000 & \multirow{2}{*}[-0.35em]{\makecell[l]{English, French, Chinese, \\ Russian, Arabic}}\\[2pt]
\cmidrule{1-3}
\textbf{Test} & TBD & TBD & \\
\bottomrule
\end{tabular}
\caption{SemEval-2021 Task 2 Data Used in the Project}
\label{tab:table1}
\end{table*}\\
For each sentence pair, the goal of the WSD system is to classify whether the target ambiguous word is used in the same meaning (T) or in different meanings (F). Below are two examples:\\
\begin{equation}
\label{eqn:eqn1}
\begin{aligned}
&\text{La \emph{souris} mange le fromage}\\
&\text{Le chat court après la \emph{souris}}
\end{aligned}
\end{equation}
\begin{center}
\rule{150pt}{1pt}
\end{center}
\begin{equation}
\label{eqn:eqn2}
\begin{aligned}
&\text{Click the right \emph{mouse} button}\\
&\text{Le chat court après la \emph{souris}}
\end{aligned}
\end{equation}
The correct label of sentence pair (\ref{eqn:eqn1}) is T since the target word \emph{souris} is used in the same meaning in both sentences, and the correct label of sentence pair (\ref{eqn:eqn2}) is F since the target word \emph{mouse} and its corresponding translation into French are used in two distinct meanings. All sentence pairs are manually annotated.\\
In the remaining part of this section, I will briefly describe how each sentence pair is \textbf{feature-engineered} (i.e. pre-processed) so that they can be used for training and testing the classifier. The \textbf{feature engineering} process is manually designed to extract features that are believed to improve the performance on WSD task. The feature engineering process involves three steps: (1) \textbf{Embedding}, (2) \textbf{Parsing}, and (3) \textbf{Concatenating}. Steps 1 and 2 are performed in parallel, and step 3 uses the results from the previous two steps to produce the final embedding that is the input of the classifier. All sentences are tokenized before being passed into steps 1 and 2.

\subsection{Step 1: BERT Embedding}
The cased DistilBERT base multilingual (mDstilBERT) model is used in this project, which is a distilled version of the BERT base multilingual model but with fewer parameters and is therefore faster and lighter than the original BERT \cite{sanh2020distilbert}. For each sentence pair in the datasets, the mDstilBERT model is applied to both sentences to analyse their full contexts and to assign a contextualised embedding to each word.\\
Because the fixed BERT vocabulary contains only around 30k tokens, any word that doesn't exist in the vocabulary will be split into smaller subwords. For example, 'milktea' is a word that doesn't exist in the BERT vocabulary, therefore when embedding this word, BERT will split it into two subwords, 'milk' and 'tea', that can be found in its dictionary, and mark all subsequent subword(s) with '\#\#', so the word 'milktea' will finally obtain two BERT embeddings, one for 'milk' and the other for '\#\#tea'. In this project, a word embedding is obtained by taking the average of its subword embeddings, for example, the BERT embedding of 'milktea' will be the averaged embedding of 'milk' and '\#\#tea'.

\subsection{Step 2: Dependency Parsing}
In parallel to the embedding step, each sentence from a sentence pair is parsed and assigned a dependency tree. In this project, spaCy\footnote{\url{https://spacy.io/}} is used for dependency parsing, which supports various languages including English, French and Chinese. The parsed results are saved in the CoNLL-U format\footnote{\url{https://universaldependencies.org/format.html}}.

\subsection{Step 3: Concatenation}
The concatenation step consists of two sub-steps: concatenation inside a sentence (step 3.1) and concatenation of two sentences in a sentence pair (step 3.2).

\subsubsection{Step 3.1: Concatenation inside a sentence}
After completing the previous two steps, each sentence is assigned a dependency tree, and each word from the sentence is assigned a contextualised embedding. The next step is to utilise information from the previous two steps to retrieve and concatenate \textbf{three embeddings} inside the sentence: the embedding of the target ambiguous word, the embedding of its \emph{head} word, and the embedding of its \emph{dependent} word(s). Note that since in a dependency tree a word can have zero or more than one dependent word(s) and/or zero head word, therefore when retrieving word embeddings, if there exist more that one dependent word for a target ambiguous word, the embeddings of the several dependent words will be addressed in two different ways: \textbf{average} (referred to as 'averaged concatenated embeddings') or \textbf{summation} (referred to as 'summed concatenated embeddings'); if the target ambiguous word doesn't have a head or dependent word, a \textbf{zero} embedding will be used instead. Each word embedding generated by the mDstilBERT model is of size 768, and therefore the concatenated word embedding, consisting of 3 BERT embeddings, is of size 2,304. When running on CPU, pre-processing one sentence pair takes around 100 seconds, and in total it takes around 200 hours to pre-process the whole training set (8,000 sentence pairs).

\subsubsection{Step 3.2: Concatenation of two sentences in a sentence pair}
\begin{figure*}[h]
  \includegraphics[width=\textwidth, height=10cm]{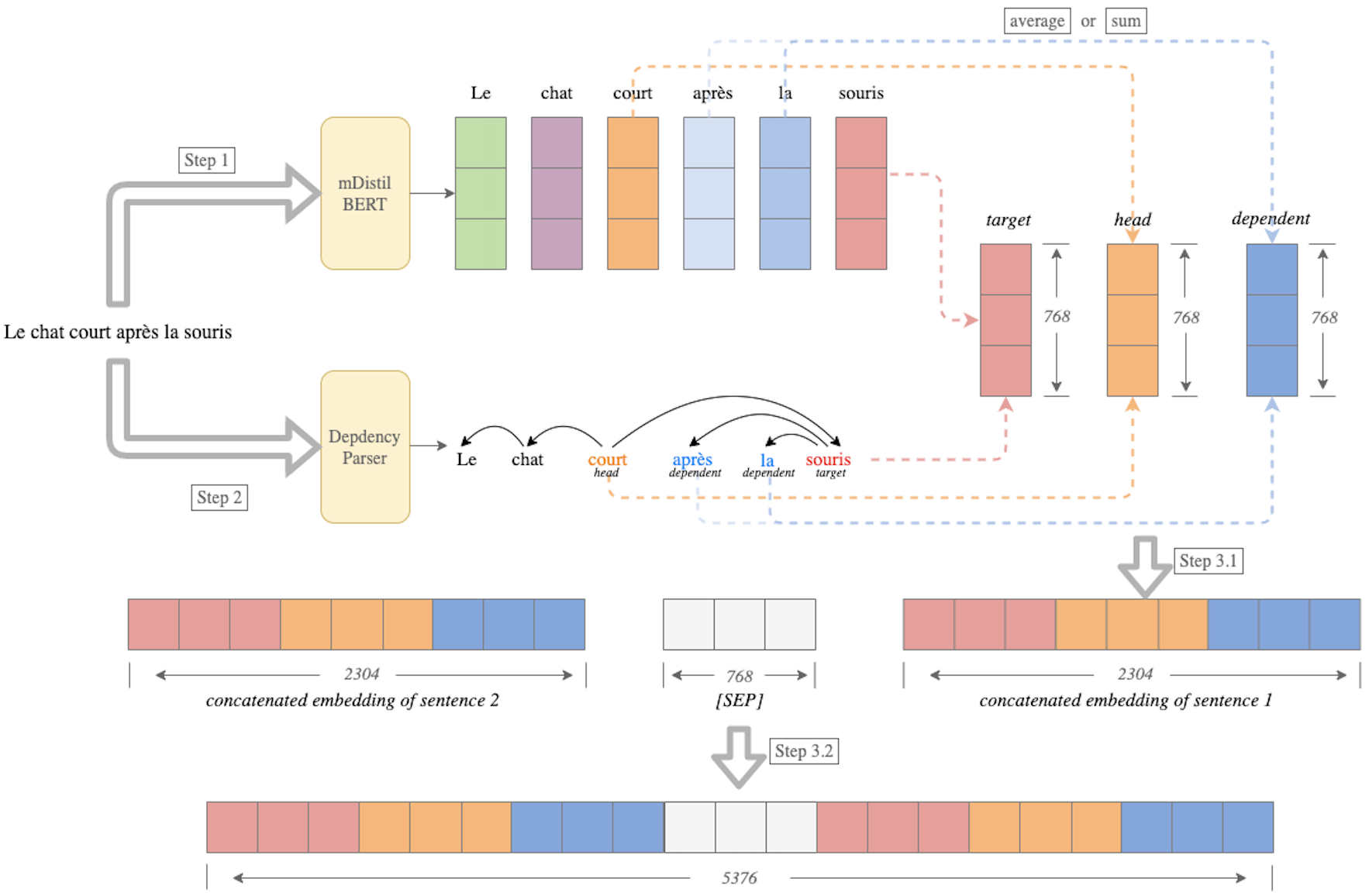}
  \caption{Pro-processing of an example sentence pair, which contains the sentence 'Le chat court après la souris'}
  \label{fig:preprocess}
\end{figure*}

After step 3.1, each sentence is assigned an embedding of size 2,304. The next step is to further concatenate the embeddings of the two sentences in a sentence pair. Because BERT was pre-trained in the form of \texttt{[CLS] sentence 1 [SEP] sentence 2 [SEP]}, and the \texttt{[SEP]} token is used to help the model understand where the previous sentence ends and where the following sentence begins. Therefore, in this project, the \texttt{[SEP]} token embedding is used to mark the boundary between the embeddings from the two sentences in a sentence pair. Since each sentence embedding is of size 2,304, and the \texttt{[SEP]} token embedding is of size 768, therefore, after concatenation, the embedding of the whole sentence pair is of size 5,376 (= 2,304$\times$2+768). Steps 1, 2, 3.1 and 3.2 are illustrated in Figure \ref{fig:preprocess}.

\section{Experiment}
Different experiments are then carried out to compare the WSD performance based on the concatenated embeddings against the baseline embeddings. The \textbf{baseline} embeddings are simply the BERT embeddings of the target ambiguous words, without taking into consideration the syntactic structure of the sentence. Therefore, the baseline embedding of each sentence is simply the BERT embedding of its target ambiguous word (of size 768), and the baseline embedding of the sentence pair is obtained by concatenating the two target embeddings using a \texttt{[SEP]} token embedding, resulting in an embedding of size 2304. \textbf{Note that compared to the baseline embeddings, the concatenated embeddings have a much larger size}.

\subsection{Logistic Regression Classifier}
\textbf{Logistic regression} is a simple but effective classification algorithm for binary classification tasks. Therefore, in this project, the logistic regression model is chosen as the first candidate for the classifier.

\subsubsection{Logistic regression with \texttt{[SEP]} as boundary marker}
First, three logistic regression classifiers are trained on the 8,000 training sentence pairs using three types of inputs: baseline embeddings, summed concatenated embeddings, and averaged concatenated embeddings. The 'liblinear' solver and L2 regularization are adopted by the classifiers. Table \ref{tab:table2} shows the test accuracies evaluated on the English development set, together with the corresponding embedding sizes.
\begin{table}[h]
\centering
\begin{tabular}{l l l}
\toprule
\textbf{Embedding} & \textbf{Embed. Size} & \textbf{Test Acc.} \\
\midrule[2pt]
\textbf{Baseline} & 2,304 & \textbf{57.13\%}\\[3pt]
\midrule
\textbf{Sum} & 5,376 & 53.69\%\\[3pt]
\midrule
\textbf{Average} & 5,376 & 53.69\%\\[3pt]
\bottomrule
\end{tabular}
\caption{Logistic regression classification using different embeddings with \texttt{[SEP]} as boundary marker}
\label{tab:table2}
\end{table}\\
As shown in Table \ref{tab:table2}, the summed and averaged concatenated embeddings don't outperform the baseline embeddings when using the logistic regression classifier with \texttt{[SEP]} token embedding as boundary marker. What's more, the summed and averaged concatenated embeddings have the same test accuracy, implying that whether summing or averaging the BERT embeddings doesn't affect the performance of the logistic regression classifier on the WSD task.

\subsubsection{Logistic regression without boundary marker}
Since the logistic regression model is a very simple model, when the input size is very large, the model performance degrades. Therefore, in order to mitigate the effect of input size on the model performance, the \texttt{[SEP]} token embedding is removed (i.e. no boundary marker), and three logistic regression classifiers are trained on the 8,000 training sentence pairs and evaluated on the English development set, which is shown in Table \ref{tab:table3}.
\begin{table}[h]
\centering
\begin{tabular}{l l l}
\toprule
\textbf{Embedding} & \textbf{Embed. Size} & \textbf{Test Acc.} \\
\midrule[2pt]
\textbf{Baseline} & 1,536 & \textbf{57.31\%}\\[3pt]
\midrule
\textbf{Sum} & 4,608 & 53.69\%\\[3pt]
\midrule
\textbf{Average} & 4,608 & 53.69\%\\[3pt]
\bottomrule
\end{tabular}
\caption{Logistic regression classification using different embeddings without boundary marker}
\label{tab:table3}
\end{table}\\
As shown in Table \ref{tab:table3}, after removing the \texttt{[SEP]} token embedding, the baseline embedding size is reduced from 2,304 to 1,536, and its test accuracy improves by 0.18\%; the concatenated embedding size is reduced from 5,376 to 4,608, however, the test accuracy doesn't improve. Therefore, when the input size is very large (i.e. concatenated embeddings), the model performance is largely dominated by the input size, and removing a single token embedding cannot directly improve the model performance. However, when the input size is moderate (i.e. baseline embeddings), removing a single token embedding significantly reduces the input size, and thus improves the model performance. This observation also supports the assumption that the logistic regression classifier performance degrades when the input size becomes very large.

\subsubsection{Logistic regression classification with different boundary markers}
If a large input size affects the model performance, and removing the \texttt{[SEP]} token embedding, which reduces the input size, can improve the model accuracy, is there an alternative method that can be used to mark the sentence boundary without significantly increasing the input size?\\
To explore the effects of different boundary markers (with different sizes) on the model performance, Table \ref{tab:table4} compares the test accuracies of the logistic regression classifiers trained using baseline embeddings with three different boundary markers: \texttt{[SEP]} token embedding, no boundary marker, and a single large integer number 9999.
\begin{table}[h]
\centering
\begin{tabular}{l l l}
\toprule
\textbf{Boundary marker} & \textbf{\makecell{Baseline \\Embed. Size}} & \textbf{Test Acc.} \\
\midrule[2pt]
\textbf{\texttt{[SEP]}} & 2,304 & 57.13\%\\[3pt]
\midrule
\textbf{None} & 1,536 & 57.31\%\\[3pt]
\midrule
\textbf{9999} & 1,537 & \textbf{58.44\%}\\[3pt]
\bottomrule
\end{tabular}
\caption{Logistic regression classification using baseline embeddings with different boundary markers}
\label{tab:table4}
\end{table}\\
As shown in Table \ref{tab:table4}, the input embeddings with no boundary marker achieve higher accuracy compared to using the \texttt{[SEP]} token embedding as boundary marker, which is because the \texttt{[SEP]} token embedding significantly increases the input size. However, the input embeddings using the single large integer number 9999 as boundary marker, which doesn't significantly increase the input size, achieve the highest accuracy.

\subsection{Multilayer Perceptron (MLP) Classifier}
The Multilayer Perceptron (MLP) model is a class of feedforward artificial neural networks. It is more sophisticated and complex than a logistic regression model and thus can learn from more complicated data patterns. Therefore, in this project, the MLP model is chosen as the second candidate for the classifier.\\
Because it's investigated in the previous section that the summed concatenated embeddings and the averaged concatenated embeddings are in fact equivalent for the classification task, therefore in this section only the summed concatenated embeddings are compared against the baseline embeddings.\\
Similar to the work by Du et al. from Tsinghua University \cite{du2019using}, I use a 2-layer MLP for classification. The MLP predicts the correct label based on the calculated probability:
\begin{equation*}
\mathbf{p} = \text{softmax}(L_2(\text{ReLU}(L_1(\mathbf{e}))))
\end{equation*}
where $\mathbf{e}$ is in the input embedding, $L_i(\mathbf{x}) = \mathbf{W}_i \mathbf{x}+\mathbf{b}_i$ are fully-connected linear layers, $\mathbf{W}_1 \in \mathbb{R}^{H\times H}$ where $H$ is the input size, and $\mathbf{W}_2 \in \mathbb{R}^{2\times H}$.

\subsubsection{MLP using different embeddings with different boundary markers}
\begin{figure}[h]
  \includegraphics[width=0.5\textwidth, height=4.5cm]{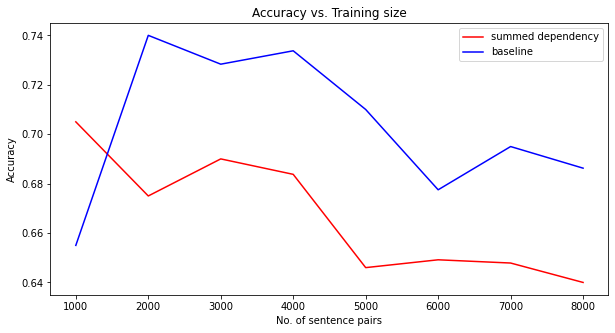}
  \caption{Test accuracy vs. Training size}
  \label{fig:training_size}
\end{figure}

Similar to the previous sections, six MLP classifiers are trained on the 8,000 training sentence pairs using two types of inputs: baseline embeddings and summed concatenated embeddings, with three different boundary markers: the \texttt{[SEP]} token embedding, no boundary marker, and the single integer number 9999. Table \ref{tab:table5} shows the test accuracies evaluated on the English development set.
\begin{table}[h]
\centering
\begin{tabular}{l l l l}
\toprule
\multirow{2}{*}{\textbf{Test Acc.}}& \multicolumn{3}{c}{\textbf{Boundary marker}} \\
& \textbf{\texttt{[SEP]}} & \textbf{None} & \textbf{9999} \\
\midrule[2pt]
\textbf{Baseline} & \textbf{68.33\%} & 65.88\% & 67.69\%\\[3pt]
\midrule
\textbf{Sum} & 63.06\% & \textbf{63.69\%} & 51.50\%\\[3pt]
\bottomrule
\end{tabular}
\caption{MLP classification using different embeddings with different boundary markers}
\label{tab:table5}
\end{table}\\
As shown in Table \ref{tab:table5}, the summed concatenated embeddings don't outperform the baseline embeddings when using the MLP classifier. However, when using the baseline embeddings as input, the \texttt{[SEP]} token embedding has the highest accuracy among the three boundary markers, while using the summed concatenated embeddings as input, using no boundary marker achieves the highest accuracy. This might be due to the fact that the size of the summed concatenated embeddings is very large, resulting in the \emph{curse of dimensionality}, and removing the boundary marker reduces the input size, hence improves the model performance.\\
Another interesting observation is that when the training size is small, the summed concatenated embeddings actually \emph{outperform} the baseline embeddings. Figure \ref{fig:training_size} shows how the test accuracy of the MLP classifier changes when the training size increases from 1,000 sentence pairs to 8,000 sentence pairs, either using the summed concatenated embeddings or the baseline embeddings.\\
\\
As shown in Figure \ref{fig:training_size}, when the training size is smaller than 1,500 (sentence pairs), the summed concatenated embeddings outperform the baseline embeddings; as the training size further increases, the performance of both the summed concatenated embeddings and the baseline embeddings degrades, with the performance of the summed concatenated embeddings degrading much faster.

\section{Discussion}
As discussed in the previous sections, the proposed concatenated embeddings, either summed or averaged, don't outperform the baseline embeddings in various experiments when the training size is larger than 1,500 (sentence pairs), which is likely due to the \emph{curse of dimensionality} caused by the large input size. Therefore, in this section, I discuss two methods designated to reduce the size of the concatenated embeddings. Besides, I also discuss one method designated to \emph{amplify} the effect of the target ambiguous word embedding in the concatenated embedding, i.e. making it more dominant than the head word embedding and the dependent word embedding.

\subsection{Dimensionality Reduction}
As shown earlier, the concatenated embeddings with no boundary marker have a higher accuracy than using either the \texttt{[SEP]} token embedding or the single integer 9999 as boundary marker. Therefore, in this section, the boundary marker is not included either. Two methods are proposed to reduce the size of the concatenated embeddings: (1) \textbf{Head-only}, and (2) \textbf{Element-wise multiplication}. The head-only method removes the dependent word embedding from the concatenated embedding, and only keeps the target ambiguous word embedding and the head word embedding. The element-wise multiplication method performs element-wise multiplication of the three embeddings, and results in a much smaller input embedding. The results are shown in Table \ref{tab:table6}.
\begin{table}[h]
\centering
\begin{tabular}{l l l}
\toprule
\textbf{Embedding} & \textbf{Embed. Size} & \textbf{Test Acc.} \\
\midrule[2pt]
\textbf{Concat. embedding} & 4,608 & \textbf{63.69\%}\\[3pt]
\midrule
\textbf{Head-only} & 3,072 & 62.56\%\\[3pt]
\midrule
\textbf{Element-wise mult.} & 1,536 & 48.19\%\\[3pt]
\bottomrule
\end{tabular}
\caption{MLP classification using concatenated embeddings with dimensionality reduction}
\label{tab:table6}
\end{table}\\
As shown in Table \ref{tab:table6}, neither of the two dimensionality-reduced embeddings outperform the concatenated embeddings. Therefore, simply reducing the embedding size will cause important information about the words to be lost, and degrade the model performance.

\subsection{Amplification of Target Ambiguous Word Embedding}
Another way of modifying the concatenated embeddings is to amplify the target ambiguous word embedding, as it is considered to be more important than the other two embeddings for WSD. Therefore, I multiply the target ambiguous word embedding by 2, while keeping the head word and dependent word embeddings unchanged. The results are shown in Table \ref{tab:table7}.
\begin{table}[h]
\centering
\begin{tabular}{l l l}
\toprule
\textbf{Embedding} & \textbf{Embed. Size} & \textbf{Test Acc.} \\
\midrule[2pt]
\textbf{No amplification} & 4,608 & \textbf{63.69\%}\\[3pt]
\midrule
\textbf{With amplification} & 4,608 & 63.19\%\\[3pt]
\bottomrule
\end{tabular}
\caption{MLP classification using concatenated embeddings with amplification of target ambiguous word embedding}
\label{tab:table7}
\end{table}\\
As shown in Table \ref{tab:table7}, the amplification of the target word embedding reduces the test accuracy. This might be due to the fact that the amplification method distorts the relative positions between the target word and the head and dependent words in the embedding space, hence loses information about the relationships among words.

\subsection{Cross-lingual Evaluation}
The last part of this project evaluates the MLP classifier, which has been previously trained on the 8,000 English sentence pairs using summed concatenated embeddings, on the trial English-French dataset\footnote{The test set has not been released by the task organization, therefore the trial set is used instead.}. The English-French sentence pairs in the trial set are pre-processed in the same way as the training set to generate summed concatenated embeddings. The evaluation results are shown in Table \ref{tab:table8}.
\begin{table}[h]
\centering
\begin{tabular}{l l l}
\toprule
\textbf{Dataset} & \textbf{Language pair} & \textbf{Test Acc.} \\
\midrule[2pt]
Training & English-English & \textbf{63.69\%}\\[3pt]
\midrule
Test & English-French & 37.50\%\\[3pt]
\bottomrule
\end{tabular}
\caption{MLP classifier evaluated on cross-lingual dataset}
\label{tab:table8}
\end{table}\\
As shown in Table \ref{tab:table8}, the test accuracy is only 37.5\% on the English-French trial set. It seems that the classifier trained on the English BERT embeddings doesn't generalise well to the French BERT embeddings. However, since the trial set is of a small size, the evaluation is likely to be biased. The model needs to be further evaluated when the test set is released. Future work may also investigate on the normalization of embeddings across languages so that the classifier can better generalise to different languages.

\section{Conclusion and Future Work}
This project investigates on the incorporation of syntactic information into the pre-trained BERT embeddings for the cross-lingual WSD task. While combining dependency parsing with BERT, the input size is inevitably increased, which leads to the curse of dimensionality and degrades the classification performance. This projects then suggests two preliminary methods to reduce the input size. Though the proposed concatenated embeddings did not achieve a significant improvement over the baseline embeddings, this project provides some first insights into the syntax-incorporated pre-trained word embedding model as well as its application in the WSD task. Future work might be dedicated to (1) reducing the size of the syntax-incorporated embeddings so as to boost the classification performance, and (2) including more cross-lingual data to stabilize the performance of the classifier across languages.

\newpage
\bibliographystyle{acl}
\bibliography{rd_report}

\end{document}